\documentclass[runningheads]{llncs}
\usepackage[utf8]{inputenc}
\usepackage{multirow}
\usepackage{xcolor}
\usepackage{amsmath}
\usepackage{amssymb}
\usepackage{algorithm}
\usepackage{algorithmic}

\usepackage{booktabs}
\usepackage{multirow}
\usepackage{graphicx}
\usepackage[table]{xcolor}
\usepackage{subcaption}
\usepackage{caption}

\definecolor{myblue}{RGB}{0, 0, 255}
\definecolor{myorange}{RGB}{255, 140, 0}

\usepackage{eccv}

\usepackage{eccvabbrv}

\usepackage{graphicx}
\usepackage{booktabs}

\usepackage[accsupp]{axessibility}  % Improves PDF readability for those with disabilities.

\usepackage[pagebackref,breaklinks,colorlinks,citecolor=eccvblue]{hyperref}
\usepackage{hyperref}

\usepackage{orcidlink}

\begin{document}

\title{PASDiff: Physics-Aware Semantic Guidance for Joint Real-World Low-Light Face Enhancement and Restoration} 

\titlerunning{PASDiff: Physics-Aware Semantic Diffusion}

% TODO FINAL: Replace with your author list. \thanks{Equal contribution}
\author{Yilin Ni\inst{1, \star} \and
Wenjie Li\inst{2, \star} \and
Zhengxue Wang\inst{3} \and 
Juncheng Li\inst{4} \and \\
Guangwei Gao\inst{3, \dagger}  \and
Jian Yang\inst{3}}

\authorrunning{Yilin Ni et al.}

\institute{College of Automation, Nanjing University of Posts and Telecommunications, Nanjing, China \and
%State Key Laboratory of Integrated Services Networks, Xidian University \and
School of Artificial Intelligence, Beijing University of Posts and Telecommunications, Beijing, China \and
PCA Lab, School of Computer Science and Engineering, Nanjing University of Science and Technology, Nanjing, China \and 
School of Computer Science and Technology, East China Normal University, Shanghai, China \\
\email{\{nixiaolin26,lewj2408\}@gmail.com, gwgao@njust.edu.cn}
}

\setcounter{footnote}{0}
\renewcommand{\thefootnote}{}
\footnotetext[0]{$^\star$Equal contribution.}
\footnotetext[0]{$^\dagger$Corresponding author: Guangwei Gao (\tt{gwgao@njust.edu.cn})}
\maketitle

%\begingroup
%\renewcommand{\thefootnote}{$\dagger$}
%\footnotetext{Corresponding author}
%\endgroup

\begin{abstract}

  Face images captured in real-world low light suffer multiple degradations—low illumination, blur, noise, and low visibility, etc. Existing cascaded solutions often suffer from severe error accumulation, while generic joint models lack explicit facial priors and struggle to resolve clear face structures. In this paper, we propose PASDiff, a Physics-Aware Semantic Diffusion in a training-free manner. To achieve a plausible illumination and color distribution, we leverage inverse intensity weighting and Retinex theory to introduce photometric constraints, thereby reliably recovering visibility and natural chromaticity. To faithfully reconstruct facial details, our Style-Agnostic Structural Injection (SASI) extracts structures from an off-the-shelf facial prior while filtering out its intrinsic photometric biases, seamlessly harmonizing identity features with physical constraints. Furthermore, we construct WildDark-Face, a real-world benchmark of 700 low-light facial images with complex degradations. Extensive experiments demonstrate that PASDiff significantly outperforms existing methods, achieving a superior balance among natural illumination, color recovery, and identity consistency. Code and dataset will be available at \url{https://github.com/IVIPLab/PASDiff}.
\keywords{Low-Light Enhancement \and Real-world Face Restoration \and Diffusion Models \and Training-Free Manner}
\end{abstract}

\section{Introduction}
\label{sec:intro}

Capturing high-quality facial images in real-world low-light scenarios, such as surveillance and handheld photography, remains a formidable challenge. To compensate for insufficient illumination, imaging systems are forced to utilize high ISO and prolonged exposure. This physical trade-off inevitably leads to a compound degradation, where low visibility, severe noise, and blur coexist. Such coupled impairments not only deteriorate visual quality but also severely hinder downstream applications like face recognition, which demand the recovery of faithful facial identities and expressions.

\begin{figure}[t]
    \centering
    \makebox[0.03\linewidth][c]{\raisebox{0.075\linewidth}{\rotatebox[origin=c]{90}{\scriptsize \textbf{Synthetic}}}}\hfill
    \includegraphics[width=0.132\linewidth]{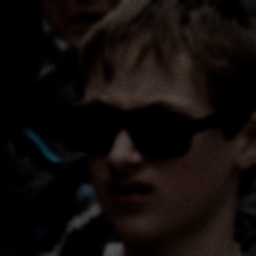}\hfill
    \includegraphics[width=0.132\linewidth]{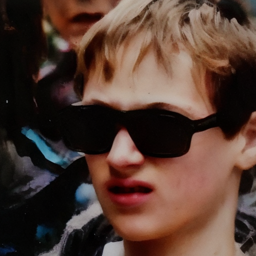}\hfill
    \includegraphics[width=0.132\linewidth]{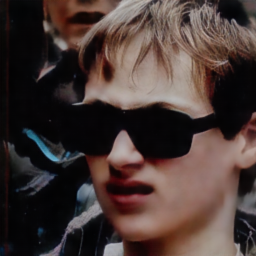}\hfill
    \includegraphics[width=0.132\linewidth]{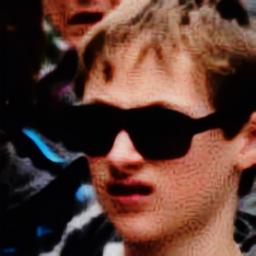}\hfill
    \includegraphics[width=0.132\linewidth]{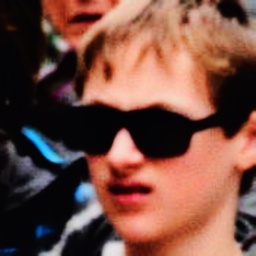}\hfill
    \includegraphics[width=0.132\linewidth]{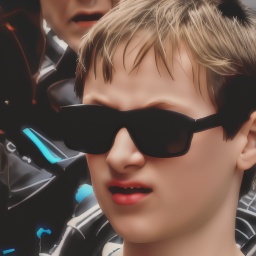}\hfill
    \includegraphics[width=0.132\linewidth]{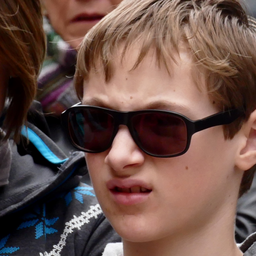}\\ 
    \vspace{0.1mm} 
    
    \makebox[0.03\linewidth][c]{\raisebox{0.075\linewidth}{\rotatebox[origin=c]{90}{\scriptsize \textbf{Real}}}}\hfill
    \includegraphics[width=0.132\linewidth]{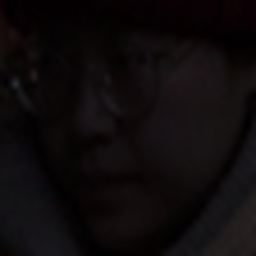}\hfill
    \includegraphics[width=0.132\linewidth]{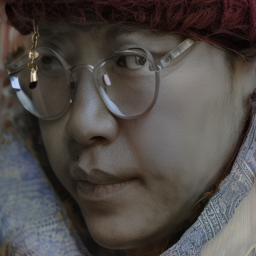}\hfill
    \includegraphics[width=0.132\linewidth]{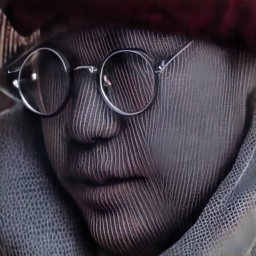}\hfill
    \includegraphics[width=0.132\linewidth]{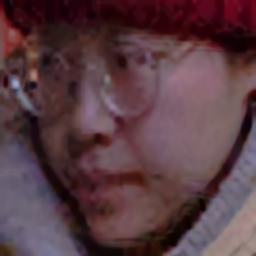}\hfill
    \includegraphics[width=0.132\linewidth]{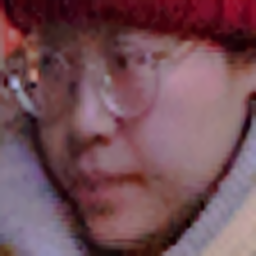}\hfill
    \includegraphics[width=0.132\linewidth]{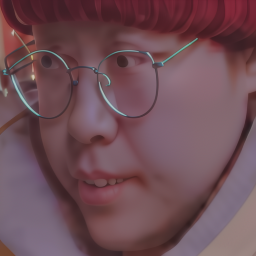}\hfill
    \includegraphics[width=0.132\linewidth]{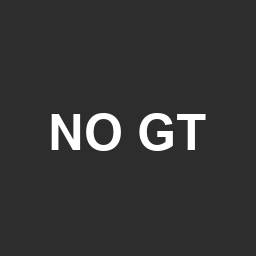}\\ 
    %\vspace{0.5mm} 
    \makebox[0.03\linewidth][c]{}\hfill
    \begin{minipage}[t]{0.132\linewidth} \centering \tiny \baselineskip=0.8\baselineskip (a) Input \end{minipage}\hfill
    \begin{minipage}[t]{0.132\linewidth} \centering \tiny \baselineskip=0.8\baselineskip (b) L-Diff$\to$\\DiffBIR \end{minipage}\hfill
    \begin{minipage}[t]{0.132\linewidth} \centering \tiny \baselineskip=0.8\baselineskip (c) DiffBIR$\to$\\L-Diff \end{minipage}\hfill
    \begin{minipage}[t]{0.132\linewidth} \centering \tiny \baselineskip=0.8\baselineskip (d) DarkIR \end{minipage}\hfill
    \begin{minipage}[t]{0.132\linewidth} \centering \tiny \baselineskip=0.8\baselineskip (e) FDN \end{minipage}\hfill
    \begin{minipage}[t]{0.132\linewidth} \centering \tiny \baselineskip=0.8\baselineskip \textbf{(f) Ours} \end{minipage}\hfill
    \begin{minipage}[t]{0.132\linewidth} \centering \tiny \baselineskip=0.8\baselineskip (g) GT \end{minipage}

    \caption{Visual comparisons on degraded synthetic and real-world data. (a) Input. Cascaded paradigms (b) L-Diff~\cite{jiang2024lightendiffusion}$\to$DiffBIR~\cite{lin2024diffbir} and (c) DiffBIR~\cite{lin2024diffbir}$\to$L-Diff~\cite{jiang2024lightendiffusion} suffer from severe noise amplification and structural collapse, respectively. Generic joint models like (d) DarkIR~\cite{feijoo2025darkir} and (e) FDN~\cite{tu2025fourier} struggle with complex degradations, leading to residual blur and detail loss. (f) Our PASDiff achieves superior perceptual quality with crisp facial details and natural illumination, effectively suppressing artifacts. (g) Ground Truth. (The real-world dataset lacks GT references.)}
    \label{fig:introduction}
\end{figure}

Previous strategies typically decompose this task into two processing tasks, i.e., Low-Light Image Enhancement (LLIE) \cite{wei2018deep, cai2023retinexformer, bai2024retinexmamba, yan2025hvi} and Blind Face Restoration (BFR) \cite{wang2021towards, li2024efficient, lin2024diffbir, wang2025osdface}. However, these methods rely on independent assumptions tailored to their specific sub-problems. Consequently, a naive cascading of these independent modules cannot solve the joint degradation. Specifically, as shown in Fig.~\ref{fig:introduction}(b) and (c), enhancing visibility first blindly amplifies latent noise, which downstream BFR models misinterpret as facial textures, leading to unnatural hallucinations. Conversely, restoring before low-light enhancing deprives BFR models of essential structural cues hidden in the darkness, resulting in irreversible over-smoothing. Alternatively, existing end-to-end approaches \cite{zhou2022lednet, zou2024vqcnir, feijoo2025darkir, tu2025fourier, liu2025liednet, xu2025urwkv} attempt to address joint degradation holistically. While achieving promising results on synthetic datasets constructed with simple noise models and uniform blur kernels, they struggle to generalize to real-world nighttime scenes. The non-linear response of sensors in extreme darkness, coupled with the intricate geometry of human faces, renders real-world degradation patterns far more complex than synthetic simulations. As observed in Fig.~\ref{fig:introduction}(d) and (e), when applied to real-world captures, these generic models often fail to recover fine-grained facial components or suffer from residual blur, highlighting the urgent need for a robust, face-specific solution.

\begin{figure*}[t]
    \centering
    \setlength{\tabcolsep}{0pt} 

    \begin{minipage}[b]{0.35\linewidth}
        \centering
        \includegraphics[width=0.235\linewidth]{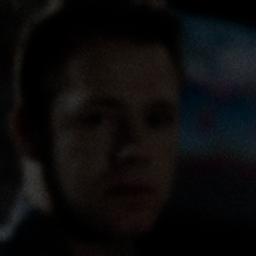}\hfill%
        \includegraphics[width=0.235\linewidth]{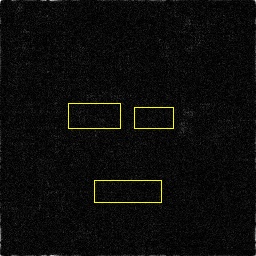}\hfill%
        \includegraphics[width=0.235\linewidth]{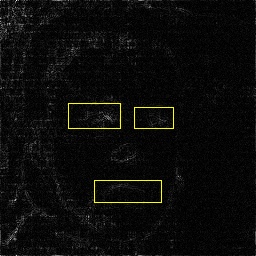}\hfill%
        \includegraphics[width=0.235\linewidth]{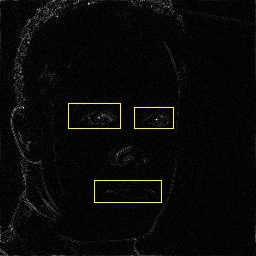}\\
        \vspace{1.5pt} 

        \includegraphics[width=0.235\linewidth]{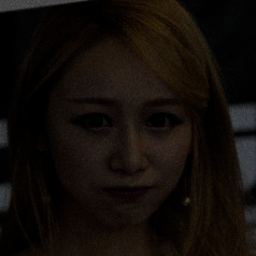}\hfill%
        \includegraphics[width=0.235\linewidth]{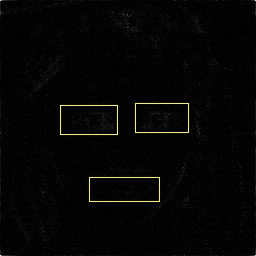}\hfill%
        \includegraphics[width=0.235\linewidth]{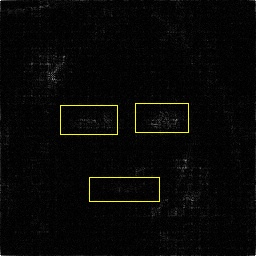}\hfill%
        \includegraphics[width=0.235\linewidth]{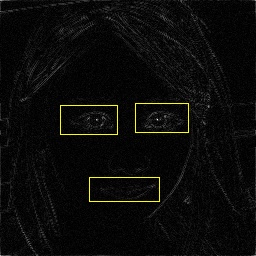}\\
        \vspace{1pt} 
        
        \begin{minipage}[t]{0.235\linewidth} \centering \tiny \baselineskip=0.8\baselineskip LQ \end{minipage}\hfill%
        \begin{minipage}[t]{0.235\linewidth} \centering \tiny \baselineskip=0.8\baselineskip Guidance\\-based \end{minipage}\hfill%
        \begin{minipage}[t]{0.235\linewidth} \centering \tiny \baselineskip=0.8\baselineskip Learning\\-based \end{minipage}\hfill%
        \begin{minipage}[t]{0.235\linewidth} \centering \tiny \baselineskip=0.8\baselineskip \textbf{Ours} \end{minipage}
    \end{minipage}%
    \hfill%
    \begin{minipage}[b]{0.18\linewidth}
        \centering
        \includegraphics[width=\linewidth]{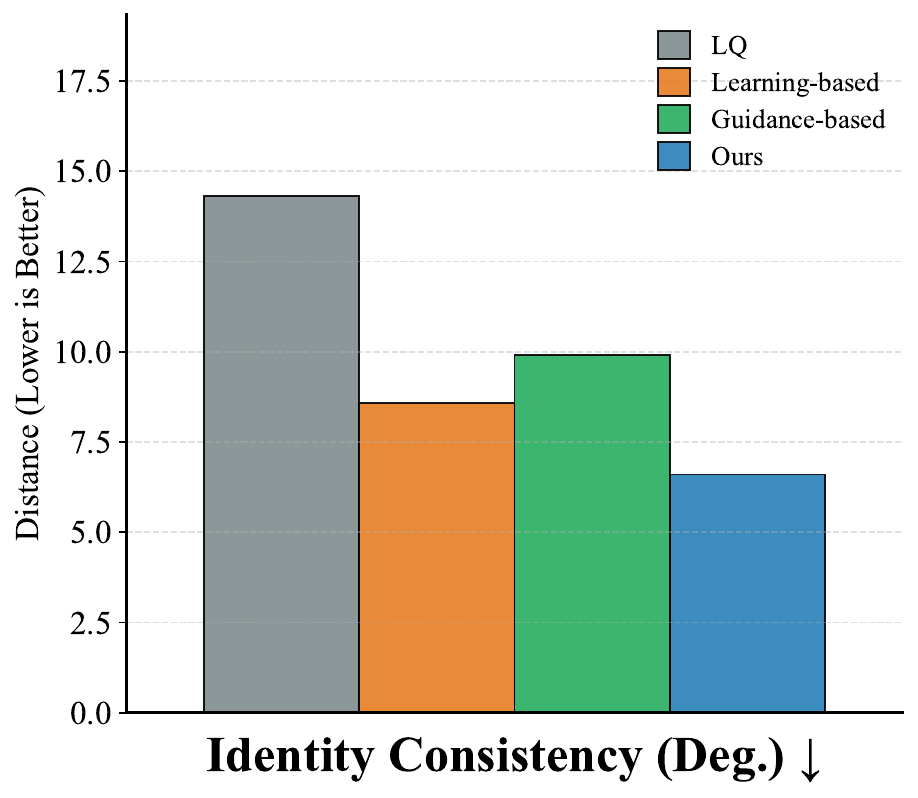}
        
        \vspace{1pt}
        \begin{minipage}[t]{\linewidth} 
            \centering \tiny \baselineskip=0.8\baselineskip Identity\\Preservation 
        \end{minipage}
    \end{minipage}%
    \hfill%
    \begin{minipage}[b]{0.45\linewidth}
        \centering
        \includegraphics[width=\linewidth]{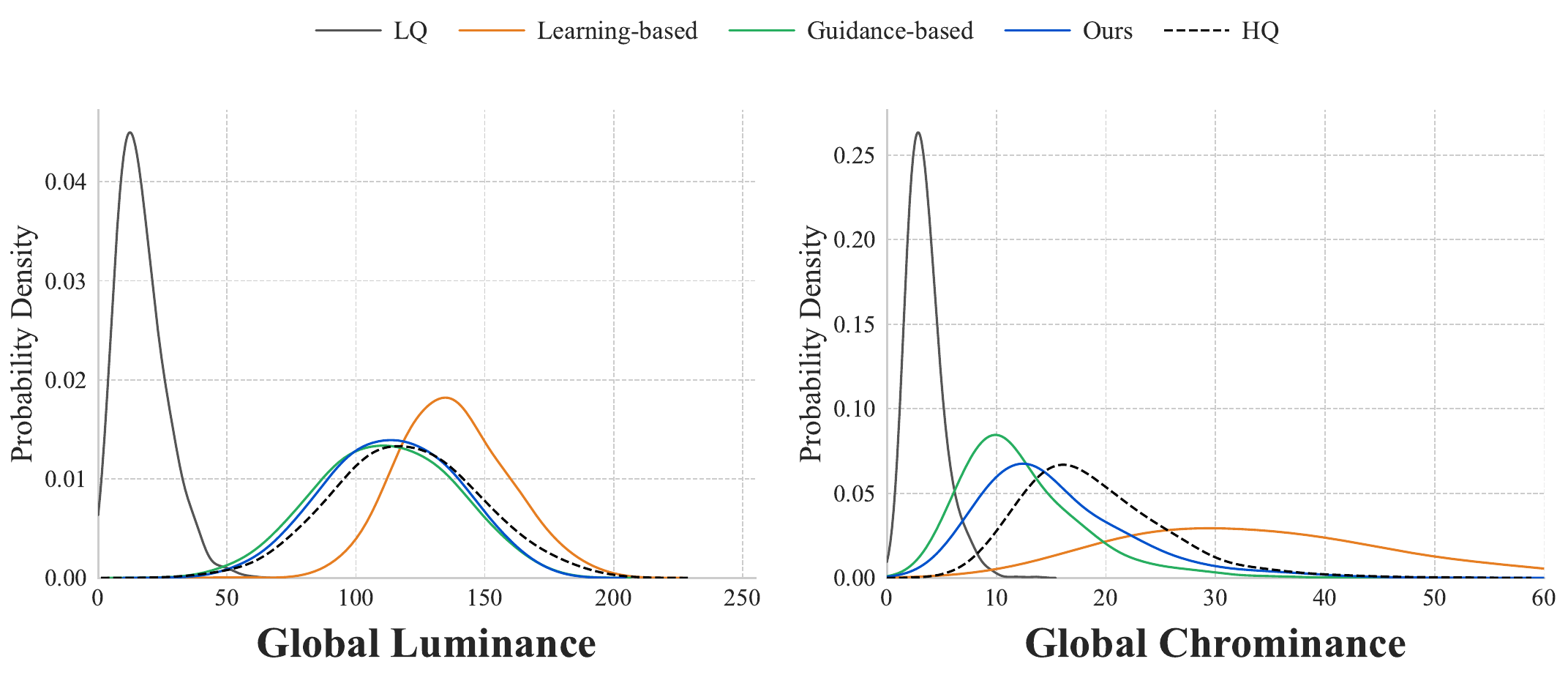}
        
        \vspace{1pt}
        \begin{minipage}[t]{\linewidth} 
            \centering \tiny \baselineskip=0.8\baselineskip Illumination Recovery and Color Restoration
        \end{minipage}
    \end{minipage}
    
    \caption{From the phase reconstructions and statistical metrics, it can be seen that end-to-end learning strategies \cite{lin2024diffbir} inherently adhere to the degraded intensity distribution, suffering from compromised facial identities and severe color shifts. Existing guidance approaches \cite{lin2025aglldiff} correct global illumination and chromaticity, but lack semantic guidance for fine geometries. In contrast, PASDiff elegantly integrates physical and structural guidance to restore crisp textures, natural illumination, and a color distribution better aligned with the high-quality reference, effectively preserving facial identity.}
    \label{fig:motivation}
    
\end{figure*}

To effectively contend with complex real-world degradations and circumvent the inherent domain gap of synthetic data-driven methods, Diffusion Probabilistic Models (DPMs) offer a promising avenue. Their robust generative priors provide the possibility of restoring high-fidelity details from severely degraded inputs. A straightforward solution for Joint Low-light Enhancement and Blind Face Restoration (Joint LL-BFR) is to directly fine-tune an existing face restoration diffusion \cite{lin2024diffbir} on paired data. However, our empirical pilot study reveals a fundamental bottleneck in this end-to-end paradigm. Under extreme low-light conditions, the network is overwhelmed by the ``dual burden'' of simultaneously rectifying drastic photometric shifts and hallucinating missing facial geometries. As shown in Fig.~\ref{fig:motivation}, the model inherently adheres to the degraded intensity distribution, either failing to reconstruct fine-grained facial structures or suffering from uncontrollable color shifts and catastrophic identity loss. In contrast, existing guidance approaches \cite{lin2025aglldiff}, which rely primarily on global physical constraints to steer the sampling, suffer from limited generative capability. Despite leveraging explicit physical priors, they lack the semantic guidance required to synthesize intricate facial geometries from noise, often producing structurally ambiguous outputs that successfully recover the low-frequency information of the image but fail to resolve fine-grained high-frequency features.

Motivated by these observations, we propose \textbf{PASDiff}, a training-free Physics-Aware Semantic Diffusion (PASDiff) that reformulates the joint task as a physically and structurally constrained generative process. We design a Multi-Objective Energy-Based Guidance strategy to explicitly steer the diffusion sampling trajectory. Specifically, our core idea is to decompose the complex degradation into distinct physical and structural components, orchestrating a synergy between the pre-trained diffusion and domain-specific priors. On the physical dimension, we incorporate spatially varying exposure constraints and Retinex-based reflectance priors, which regulate the optimization path, ensuring that both the recovered illumination and chromaticity align with natural scene statistics. On the structural dimension, to address the ill-posed nature of blind face restoration, we devise a Style-Agnostic Structural Injection (SASI) mechanism. We leverage an off-the-shelf restore to provide structural cues, but we identify that these priors carry incorrect illumination and color estimates, which can introduce incorrect global style biases into the generative process. To solve this, we propose a Statistic-Aligned Guidance Loss. By dynamically aligning the first and second-order statistics (mean and variance) of the guidance signal with the current diffusion state, this mechanism strictly distills high-frequency structural gradients from the prior while statistically filtering out its low-frequency biases in both luminance and chromaticity. This facilitates an effective decoupling of texture recovery from illumination and color enhancement, enabling our model to synthesize plausible details that are both structurally faithful and visually harmonious. Through our design, as shown in Fig.~\ref{fig:motivation}, restoration results exhibit faithful identity preservation and natural illumination distribution.

In summary, our main contributions are as follows:
\begin{itemize}
    \item[$\bullet$] We propose PASDiff, a training-free framework for joint low-light enhancement and face restoration. By reformulating the task as a dually constrained generation, we harness priors of diffusion without training and paired data.
    
    \item[$\bullet$] We devise a SASI strategy via a Statistic-Aligned Guidance Loss, which elegantly decouples texture recovery from global photometry, enabling the precise distillation of structural semantics from off-the-shelf priors while explicitly filtering out their intrinsic lighting and color biases.
    
    \item[$\bullet$] We construct WildDark-Face, a real-world benchmark comprising 700 facial images with complex compound degradations. Extensive experiments demonstrate that PASDiff surpasses existing solutions, delivering superior perceptual quality, natural illumination, and identity preservation.
\end{itemize}

\section{Related Work}

\subsection{Joint Restoration of Compound Degradation}
Restoring images captured in unconstrained environments is highly challenging. We briefly review methods addressing individual degradations and recent attempts toward joint restoration.

\noindent\textbf{Low-Light Image Enhancement.}
Early approaches relied on Histogram Equalization or Retinex theory \cite{guo2016lime} to decompose reflectance and illumination. Deep learning methods \cite{wei2018deep, wu2022uretinex} integrated these physical models into neural networks, while unsupervised approaches like Zero-DCE \cite{guo2020zero} eliminated the need for paired data. Recently, generative methods such as LLFlow \cite{wang2022low} and GDP \cite{fei2023generative} have employed normalizing flows and diffusion priors to model complex light distributions. Furthermore, DLFN \cite{zhou2026diffusion} integrated diffusion models with Laplacian decomposition to balance noise suppression and detail preservation. To improve efficiency, RetinexMamba \cite{bai2024retinexmamba} leveraged State Space Models for long-range dependency modeling, and HVI \cite{yan2025hvi} introduced a hue-value-insensitive color space to reduce distortion. However, these methods operate on global statistics and lack specific awareness of facial semantics, often leading to over-smoothing or color artifacts on human faces.

\noindent\textbf{Blind Face Restoration.}
Blind Face Restoration (BFR)~\cite{li2025survey} focuses on recovering facial details from degraded inputs. While early works used geometric priors \cite{chen2018fsrnet}, GAN-based methods like GFPGAN \cite{wang2021towards} and CodeFormer \cite{zhou2022towards} have become mainstream by leveraging pre-trained StyleGAN latent spaces or VQ-codebooks. More recently, diffusion-based models such as DiffBIR \cite{lin2024diffbir} have established strong baselines by decoupling degradation removal and generation. To accelerate inference, OSDFace \cite{wang2025osdface} utilized a vector-quantized dictionary to achieve one-step restoration. Nevertheless, these models assume "normal" lighting; severe noise and uneven illumination in low-light conditions disrupt their feature extraction, causing structural deformation.

\noindent\textbf{Restoration of Coupled Degradations.}
Addressing multiple degradations simultaneously is critical but difficult. Simple cascaded strategies often lead to error accumulation. Pioneering this direction, LEDNet \cite{zhou2022lednet} proposed a specific encoder-decoder architecture with a re-weighting mechanism to jointly handle low-light and blur. Unified networks like AirNet \cite{li2022all} and PromptIR \cite{potlapalli2023promptir} further attempted to handle various degradations within a single framework. Most recently, InstructIR \cite{conde2024instructir} introduced instruction tuning to guide the model in removing specific degradation types. Despite these efforts, methods like DarkIR \cite{feijoo2025darkir} and FDN \cite{tu2025fourier}, which utilized spatial-frequency attention or Fourier decoupling, usually operate discriminatively. They struggle to hallucinate high-frequency details needed for face restoration and often rely on specific blur assumptions, lacking the generalization ability of generative priors.

\subsection{Diffusion Models for Inverse Problems}
Leveraging pre-trained diffusion priors for inverse problems has gained significant traction in low-level vision tasks. One type of work, like DPS \cite{chung2022diffusion}, DDRM \cite{kawar2022denoising}, and MCS \cite{li2025measurement}, approximated the posterior distribution using measurement errors or singular value decomposition, while DDNM \cite{wang2023ddnm} introduced null-space decomposition for linear consistency. For unknown degradations, BlindDPS \cite{chung2023parallel} jointly modeled the image and degradation operator. FreeDoM \cite{yu2023freedom} employed multiple energy functions to guide sampling without retraining, DDPG \cite{garber2024image} proposed an iteratively preconditioned strategy to robustly solve non-linear inverse problems, and SSDiff \cite{li2025self} constructed pseudo-labels to assist the guidance. Recent advancements like DAPS \cite{zhang2025improving} and FlowDPS \cite{kim2025flowdps} further improved sampling stability and generative quality.

In the realm of illumination and visibility enhancement, methods adapt these priors using unpaired data \cite{lan2025exploiting} or by learning specific degradation representations \cite{wang2025lldiffusion}. Notably, Reti-Diff \cite{he2025reti} integrated Retinex theory with latent diffusion to explicitly tackle illumination degradation. Crucially, however, these guidance strategies primarily focus on handling single degradation types independently. They lack the mechanism to seamlessly incorporate the composite physical constraints required for handling complex coupled degradations (e.g., simultaneous low-light enhancement and identity-preserving face restoration). In this work, we bridge this gap by designing a multi-objective energy function to guide the diffusion trajectory toward a high-fidelity, well-lit facial manifold.

\section{Methodology}

\subsection{Preliminaries}
\label{sec:preliminaries}

\noindent\textbf{Denoising Diffusion Probabilistic Models (DDPM).} 
DDPM \cite{ho2020denoising} defines a forward diffusion process that gradually adds Gaussian noise to a data sample $x_0 \sim q(x_0)$ over $T$ steps. The forward transition $q(x_t|x_{t-1})$ is parameterized as a Markov chain:
\begin{equation}
    q(x_t|x_{t-1}) = \mathcal{N}(x_t; \sqrt{1-\beta_t}x_{t-1}, \beta_t \mathbf{I}),
\end{equation}
where $\beta_t$ is a pre-defined variance schedule. Let $\alpha_t = 1-\beta_t$ and $\bar{\alpha}_t = \prod_{i=1}^t \alpha_i$. A remarkable property of this process is that we can sample $x_t$ at any arbitrary timestep $t$ directly from $x_0$:
\begin{equation}
    x_t = \sqrt{\bar{\alpha}_t}x_0 + \sqrt{1-\bar{\alpha}_t}\epsilon, \quad \epsilon \sim \mathcal{N}(0, \mathbf{I}).
\end{equation}
The reverse process learns to denoise $x_t$ to recover $x_0$. It is modeled by a neural network $\epsilon_\theta(x_t, t)$ which predicts the noise component added in the forward process. The reverse transition $p_\theta(x_{t-1}|x_t)$ is defined as $p_\theta(x_{t-1}|x_t) = \mathcal{N}(x_{t-1}; \mu_\theta(x_t, t), \Sigma_\theta(x_t, t))$, where the mean $\mu_\theta(x_t, t)$ is derived as:
\begin{equation}
    \mu_\theta(x_t, t) = \frac{1}{\sqrt{\alpha_t}} \left( x_t - \frac{\beta_t}{\sqrt{1-\bar{\alpha}_t}} \epsilon_\theta(x_t, t) \right).
\end{equation}
In practice, we can directly estimate $x_0$ from the predicted noise $\epsilon_\theta(x_t, t)$ via:
\begin{equation}
\hat{x}_0 = \frac{x_t - \sqrt{1-\bar{\alpha}t}\epsilon\theta(x_t, t)}{\sqrt{\bar{\alpha}_t}}.
\label{eq:x0_pred}
\end{equation}

\noindent\textbf{Classifier Guidance.} 
To introduce semantic control into the generation process, Dhariwal et al. \cite{dhariwal2021diffusion} proposed classifier guidance. Instead of training a conditional diffusion model from scratch, this approach modifies the sampling trajectory of a pre-trained unconditional model using gradients from an auxiliary classifier $c_\phi(y|x_t)$. The perturbed reverse transition probability can be approximated as a Gaussian distribution with a shifted mean:
\begin{equation}
    p_{\theta,\phi}(x_{t-1}|x_t, y) \approx \mathcal{N}(x_{t-1}; \mu_\theta(x_t, t) + \Sigma_\theta(x_t, t) \nabla_{x_t} \log c_\phi(y|x_t), \Sigma_\theta(x_t, t)),
    \label{eq:classifier_guidance}
\end{equation}
where $s$ is the guidance scale. The gradient term $g = \nabla_{x_t} \log c_\phi(y|x_t)$ acts as a guidance signal that biases the sampling distribution toward the target semantics defined by $y$, effectively steering the generative process without altering the pre-trained weights.

\begin{figure*}[t]
    \centering
    \includegraphics[width=\textwidth]{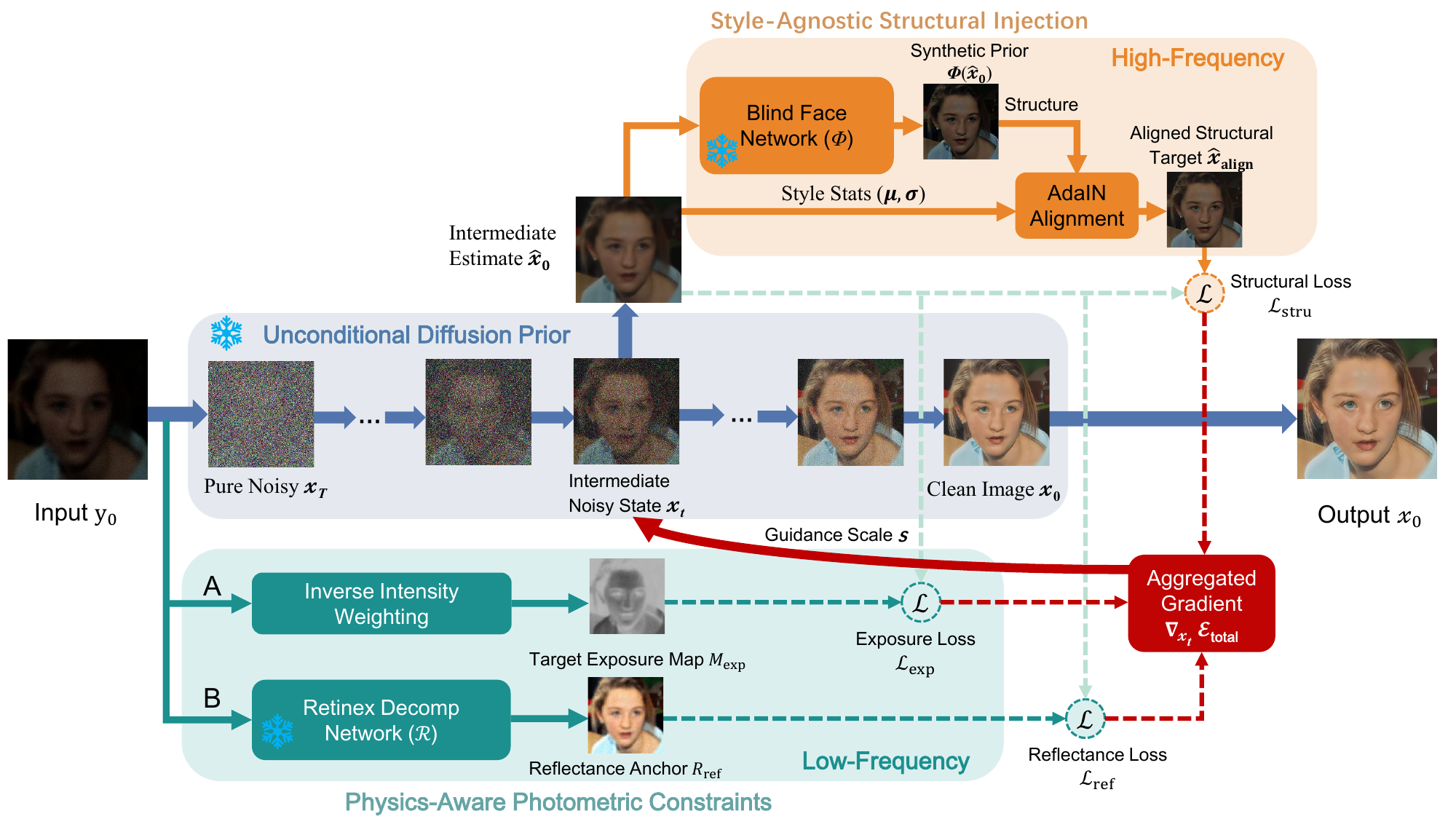}
   
    \caption{\textbf{Overall framework of the proposed training-free PASDiff.} Our method reformulates joint restoration via dual-dimensional diffusion guidance. Physically, Retinex-based photometric constraints steer the sampling trajectory to recover natural illumination and chromaticity. Structurally, our Style-Agnostic Structural Injection (SASI) and Statistic-Aligned Guidance Loss extract high-frequency facial semantics from priors while explicitly filtering out their intrinsic lighting biases. This synergy seamlessly harmonizes texture realism with physical reliability.}
    \label{fig:framework}
\end{figure*}

\subsection{Physics-Aware Semantic Guidance Framework}
\label{sec:framework}

As analyzed in Sec.~\ref{sec:intro}, existing solutions for complex degraded low-light face images face a fundamental conflict between \textit{structural fidelity} and \textit{photometric restoration}. To address this conflict, we propose PASDiff, a train-free framework that decouples this task into two orthogonal objectives: \textit{photometric correction} and \textit{structural refinement}, unifying them within the latent trajectory of an unconditional DDPM. As illustrated in Fig.~\ref{fig:framework} and Algorithm~\ref{alg:inference}, given a degraded input $y_0$, our goal is to sample $x_0$ from a posterior distribution $p(x_0 \mid y_0)$. Since the exact posterior is intractable, we approximate it by steering the reverse process of an unconditional DDPM using a composite energy function $\mathcal{E}_{total}$. Specifically, at each timestep $t$, we estimate the clean image $\hat{x}_0$ from the current noisy state $x_t$. Then, we apply physical constraints to correct the illumination and color of $\hat{x}_0$, while leveraging a Style-Agnostic Structural Injection (SASI) to extract high-frequency facial semantics without introducing incorrect global style biases. The aggregated gradient $\nabla_{x_t} \mathcal{E}_{total}$ ultimately guides the trajectory toward a high-fidelity manifold. In the following, we elaborate on its two key components: Physics-Aware Photometric Constraints and Style-Agnostic Structural Injection.

\begin{algorithm}[t]
\caption{Inference via Physics-Aware Semantic Guidance}
\label{alg:inference}
\begin{algorithmic}[1]
\REQUIRE Pre-trained diffusion model $(\mu_\theta(x_t, t), \Sigma_\theta(x_t, t))$, Restoration Model $\Phi$, Retinex Net $\mathcal{R}$, guidance scale $s$, gradient steps $N$.
\STATE \textbf{Input:} A low-light face image $y_0$ with complex degradation.
\STATE \textbf{Output:} High-quality normal-light face image $\hat{x}_0$.

\STATE $M_{exp} \leftarrow \text{GetExposure}(y_0); \quad R_{ref} \leftarrow \mathcal{R}(y_0)$
\STATE $x_T \leftarrow \sqrt{\bar{\alpha}_T}y_0 + \sqrt{1-\bar{\alpha}_T}\epsilon, \quad \epsilon \sim \mathcal{N}(0, \mathbf{I})$

\FOR{$t = T$ \textbf{to} $1$}
    \STATE $\mu_t, \Sigma_t \leftarrow \mu_\theta(x_t, t), \Sigma_\theta(x_t, t)$
    \STATE $\hat{x}_0 \leftarrow \frac{1}{\sqrt{\bar{\alpha}_t}} x_t - \frac{\sqrt{1-\bar{\alpha}_t}}{\sqrt{\bar{\alpha}_t}} \epsilon_\theta(x_t, t)$
    
    \REPEAT
        \STATE $\mathcal{L}_{phy} = \lambda_{exp}\| \text{Mean}_c(\hat{x}_0) - M_{exp} \|^2 + \lambda_{ref}\| \mathcal{R}(\hat{x}_0) - R_{ref} \|^2; \quad \mathcal{L}_{stru} = \lambda_{stru}\| \hat{x}_0 - \text{AdaIN}(\Phi(\hat{x}_0), \hat{x}_0) \|^2$
        \STATE $g \leftarrow \nabla_{\hat{x}_0} (\mathcal{L}_{phy} + \mathcal{L}_{stru})$
        \STATE $x_t \sim \mathcal{N}(\mu_t - s \Sigma_t g, \Sigma_t)$
        \STATE $\hat{x}_0 \leftarrow \frac{1}{\sqrt{\bar{\alpha}_t}} x_t - \frac{\sqrt{1-\bar{\alpha}_t}}{\sqrt{\bar{\alpha}_t}} \epsilon_\theta(x_t, t)$
    \UNTIL{$N-1$ times}
    
    \STATE $x_{t-1} \sim \mathcal{N}(\mu_t - s \Sigma_t g, \Sigma_t)$
\ENDFOR
\RETURN $\hat{x}_0$
\end{algorithmic}
\end{algorithm}

\noindent\textbf{Physics-Aware Photometric Constraints.}
While generative models provide powerful priors, they lack explicit knowledge of the scene's physical lighting conditions. In low-light scenarios, unconstrained generation often leads to inconsistent chromatic shifts, as the model attempts to synthesize chromatic features from noise without a reliable reference. To ground the generative process in physical reality, we leverage Retinex theory~\cite{land1977retinex} as the theoretical foundation for our dual constraints. Retinex theory decomposes an image $I$ into an illumination component $L$ (governing visibility) and a reflectance component $R$ (governing intrinsic color), modeled as $I = R \circ L$. Based on this physical decoupling, we guide the diffusion trajectory via two complementary terms: one regulating the illumination $L$ to ensure adequate visibility, and the other constraining the reflectance $R$ to promote natural chromaticity.

\begin{figure*}[t]
    \centering
    \begin{minipage}[t]{0.38\textwidth}
        \vspace{0pt}
        \captionof{table}{Face recognition accuracy comparisons.}
        \label{tab:face_acc}
        \resizebox{\textwidth}{!}{
        \begin{tabular}{l|c}
            \toprule
            Method & Accuracy $\uparrow$ \\
            \midrule
            L-Diff~\cite{jiang2024lightendiffusion} $\rightarrow$ TSD-SR~\cite{dong2025tsd} & 60.06\% \\
            L-Diff~\cite{jiang2024lightendiffusion} $\rightarrow$ DiffBIR~\cite{lin2024diffbir} & 63.96\% \\
            TSD-SR~\cite{dong2025tsd} $\rightarrow$ L-Diff~\cite{jiang2024lightendiffusion} & 60.71\% \\
            DiffBIR~\cite{lin2024diffbir} $\rightarrow$ L-Diff~\cite{jiang2024lightendiffusion} & 49.68\% \\
            DarkIR~\cite{feijoo2025darkir} & 58.44\% \\
            LIEDNet~\cite{liu2025liednet} & 57.14\% \\
            LEDNet~\cite{zhou2022lednet} & 57.14\% \\
            VQCNIR~\cite{zou2024vqcnir} & 49.68\% \\
            URWKV~\cite{xu2025urwkv} & 51.95\% \\
            FDN~\cite{tu2025fourier} & 61.04\% \\
            \midrule
            \textbf{PASDiff (Ours)} & \textbf{71.43\%} \\
            \bottomrule
        \end{tabular}
        }
    \end{minipage}
    \hfill
    \begin{minipage}[t]{0.58\textwidth}
        \vspace{4mm}
        \begin{minipage}[t]{0.48\textwidth}
            \centering
            \includegraphics[width=\textwidth]{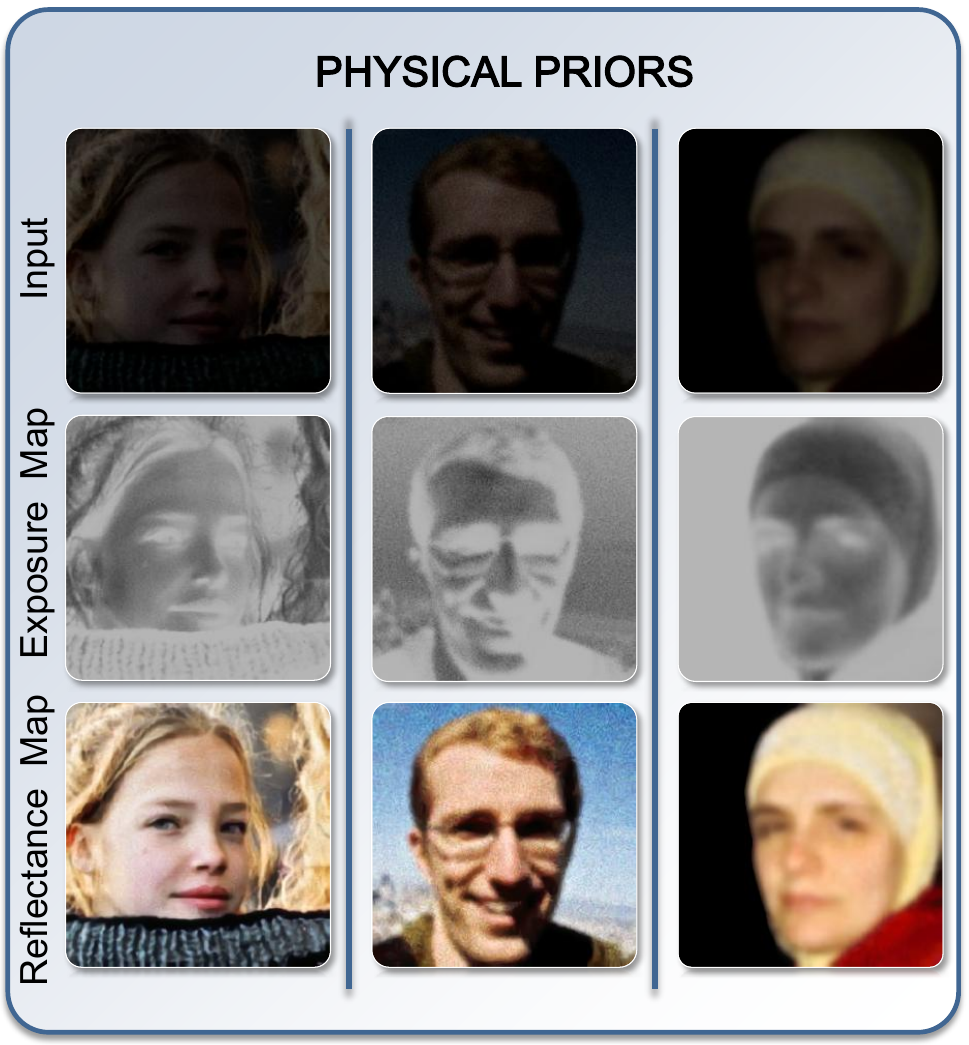}
        \end{minipage}
        \hfill
        \begin{minipage}[t]{0.48\textwidth}
            \centering
            \includegraphics[width=\textwidth]{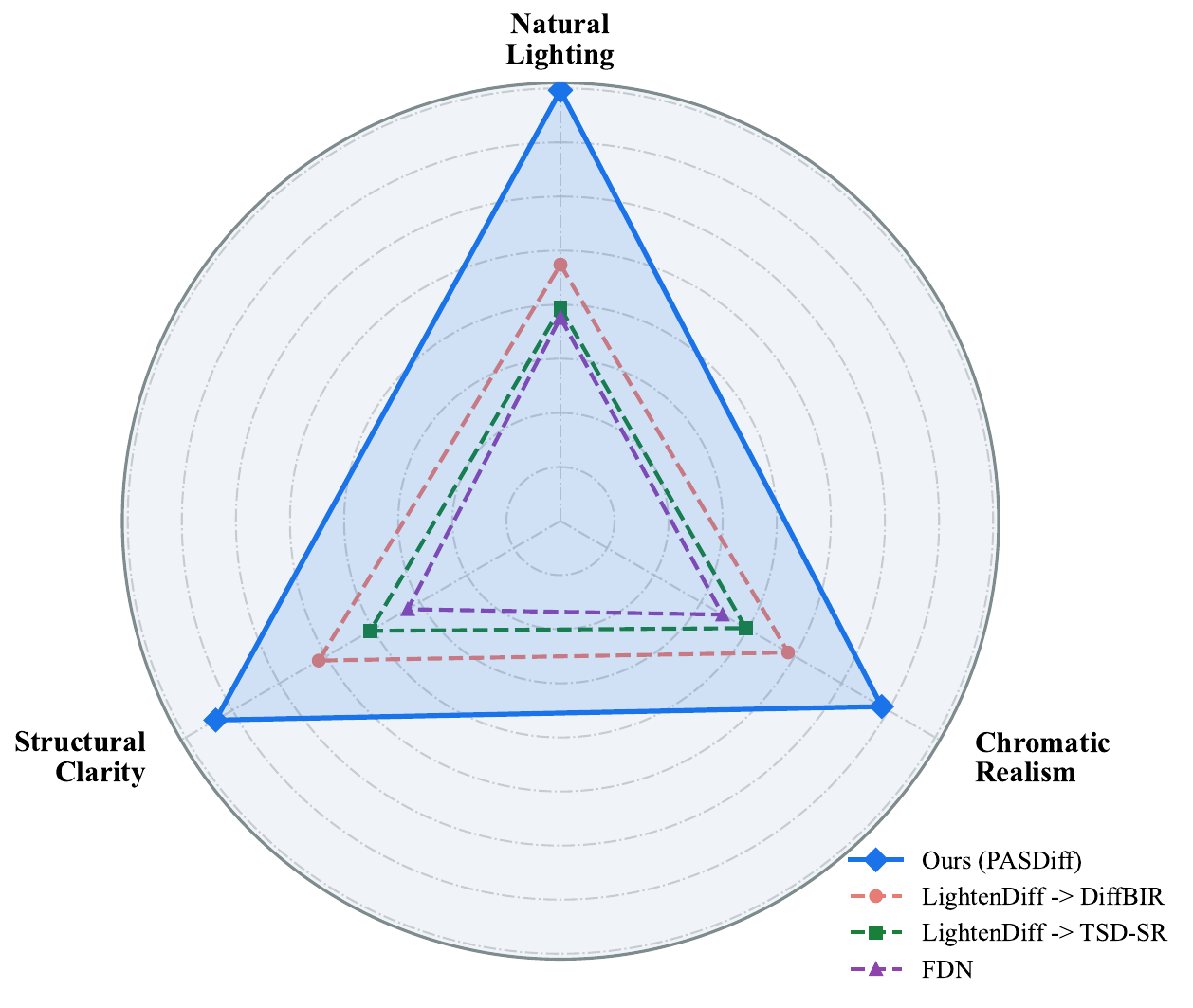}
        \end{minipage}
        
        \captionof{figure}{\textbf{Left:} Visualization of physical priors (Exposure and Reflectance maps). \textbf{Right:} Subjective preferences from the user study.}
        \label{fig:priors_user}
    \end{minipage}
\end{figure*}

\noindent\textit{(a) Spatially-Varying Exposure Guidance.}
A critical challenge in low-light enhancement is the extreme dynamic range: blindly boosting global brightness often renders dark regions visible at the cost of blowing out originally bright areas (e.g., street lamps or reflections). Uniform exposure adjustments fail to balance these conflicting demands. To address this, we formulate a target exposure map $M_{exp}$ based on an inverse intensity weighting strategy \cite{li2022cudi}. Specifically, to decouple lightness from chromaticity, we transform the input $y_0$ into the HSI color space and extract the Intensity component $I_{in}$ (calculated as the pixel-wise average of RGB channels). To construct a spatially adaptive guide, we compute the deviation of local intensity from the global average intensity $\bar{I}_{in}$. The target exposure map $M_{exp}$ (Fig.~\ref{fig:priors_user} Left) is then synthesized as:
\begin{equation}
    M_{exp} = \alpha + \beta \cdot \text{Norm}(\bar{I}_{in} - I_{in}),
\end{equation}
where $\text{Norm}(\cdot)$ denotes min-max normalization. Based on statistical observations of low-light distributions, the hyperparameters $\alpha$ (base exposure) and $\beta$ (adjustment amplitude) are empirically set to 0.46 and 0.25, respectively. This map acts as a pixel-wise attention mechanism: assigning higher target exposures to underexposed regions while restricting gains in brighter areas. Consequently, the exposure loss is defined as:
\begin{equation}
    \mathcal{L}_{exp} = \| \text{Mean}_c(\hat{x}_0) - M_{exp} \|_2^2,
\end{equation}
where $\text{Mean}_c(\cdot)$ computes the average along the channel dimension. By minimizing $\mathcal{L}_{exp}$, we enforce a physically balanced global illumination distribution on the estimated $\hat{x}_0$.

\noindent\textit{(b) Retinex-based Reflectance Prior.}
Recovering precise chromaticity from extreme darkness is an intrinsically ill-posed problem. The lack of sufficient photon counts inevitably leads to color undersaturation and unpredictable chromatic shifts. To constrain the solution space within a plausible color manifold, we exploit the illumination-invariant property of reflectance. According to the Retinex theory, the reflectance component represents the intrinsic chromatic properties of objects, and this chromatic consistency provides robust cues even under extremely low-light conditions. Therefore, we utilize the reflectance map (Fig.~\ref{fig:priors_user} Left) extracted from the input $y_0$ as a robust "chromatic anchor" to constrain the solution space of the restoration process. We employ a pre-trained Retinex decomposition network \cite{fu2023learning}, denoted as $\mathcal{R}(\cdot)$, to extract the reference reflectance $R_{ref} = \mathcal{R}(y_0)$. We premise that a valid high-quality restoration $\hat{x}_0$ should maintain chromatic consistency with the intrinsic reflectance of the input. Thus, we formulate the reflectance loss $\mathcal{L}_{ref}$ as:
\begin{equation}
    \mathcal{L}_{ref} = \| \mathcal{R}(\hat{x}_0) - R_{ref} \|_2^2.
\end{equation}
By minimizing this term, we effectively restrict the generative diversity toward a visually plausible color manifold. This ensures that the diffusion model \cite{li2026seeing} focuses its capacity on synthesizing high-frequency textures while aligning closely with the scene's available chromatic cues. The total physical guidance energy $\mathcal{L}_{phy}$ is then formulated as a weighted sum:
\begin{equation}
    \mathcal{L}_{phy} = \lambda_{exp}\mathcal{L}_{exp} + \lambda_{ref}\mathcal{L}_{ref}.
\end{equation}

\noindent\textbf{Style-Agnostic Structural Injection.}
While the aforementioned constraints ensure photometric plausibility, they operate primarily in the low-frequency domain. Recovering intricate high-frequency facial details (e.g., pores and eyelashes) from severe degradations remains a formidable challenge that physics alone cannot solve. To bridge this gap, we introduce a semantic prior from a potent blind face restoration network \cite{wang2025osdface}, denoted as the external restoration prior $\Phi$. However, integrating this prior into our framework introduces a critical domain conflict: predictions from off-the-shelf models ($\hat{x}_{prior} = \Phi(\hat{x}_0)$) are typically biased toward canonical laboratory lighting and synthetic color distributions. Directly minimizing $\|\hat{x}_0 - \hat{x}_{prior}\|_2^2$ would force the generative process to mimic these synthetic styles, which contradicts our physically grounded constraints and degrades visual naturalness. To resolve this dilemma, we propose a Style-Agnostic Structural Injection strategy. Our core insight is that the \textit{identity} and \textit{structure} of a face reside in high-frequency spatial variations, while the photometric \textit{style} is dominated by low-frequency global statistics. Therefore, our goal is to extract the structural gradients strictly from $\Phi$ while statistically stripping away its photometric biases. We achieve this by aligning the feature statistics (mean $\mu$ and standard deviation $\sigma$) of the prior prediction with the current intermediate state $\hat{x}_0$ via Adaptive Instance Normalization (AdaIN). The aligned structural target $\hat{x}_{align}$ is formulated as:
\begin{equation}
    \hat{x}_{align} = \sigma(\hat{x}_0) \left( \frac{\Phi(\hat{x}_0) - \mu(\Phi(\hat{x}_0))}{\sigma(\Phi(\hat{x}_0)) + \epsilon} \right) + \mu(\hat{x}_0),
\end{equation}
where $\epsilon$ is a small constant for numerical stability. This transformation effectively normalizes the prior's output to a zero-mean, unit-variance space, stripping away its original illumination and color biases, and then re-projects it onto the intensity distribution of $\hat{x}_0$. Consequently, $\hat{x}_{align}$ inherits the high-fidelity structural details from $\Phi$ but strictly adheres to the illumination and color atmosphere of $\hat{x}_0$ (which is governed by our physical constraints). Finally, we define the structural guidance loss $\mathcal{L}_{stru}$ as the distance between the current estimate and this aligned target:
\begin{equation}
    \mathcal{L}_{stru} = \| \hat{x}_0 - \hat{x}_{align} \|_2^2.
\end{equation}
By combining this with the physical gradients, the total guidance gradient $g_{total}$ is derived by aggregating the weighted physical and structural objectives:
\begin{equation}
    g_{total} = \nabla_{\hat{x}_0} (\mathcal{L}_{phy} + \mathcal{L}_{stru}) = \nabla_{\hat{x}_0} (\lambda_{exp}\mathcal{L}_{exp} + \lambda_{ref}\mathcal{L}_{ref} + \lambda_{stru}\mathcal{L}_{stru}),
\end{equation}
where, to handle the varying magnitudes of the loss terms, the balancing weights are empirically set to $\lambda_{exp}=1200$, $\lambda_{ref}=0.03$, and $\lambda_{stru}=10000$. This aggregated gradient steers the diffusion process toward a manifold that is both photometrically natural and structurally faithful.

\section{Experiments}

\subsection{Experimental Settings}

\noindent\textbf{Implementation Details.} 
Following previous methods~\cite{chung2022diffusion,wang2023ddnm,kim2025flowdps}, our method is built upon a pre-trained unconditional diffusion model ($256 \times 256$ resolution) trained on ImageNet~\cite{dhariwal2021diffusion}. To formulate the multi-objective guidance, we adopt a pre-trained Retinex decomposition network~\cite{fu2023learning} to provide photometric constraints and leverage an off-the-shelf blind face restoration network~\cite{wang2025osdface} to extract structural semantics. For inference, the standard diffusion process in our method consists of $T=10$ timesteps to execute noise injection and attribute guidance. All experiments are conducted on a single NVIDIA RTX 3090 GPU.

\begin{table*}[t]
\centering
\caption{Quantitative comparisons with existing methods on the synthetic FFHQ dataset and the real-world WildDark-Face benchmark. 'C' and 'J' denote Cascaded and Joint general restoration approaches, respectively. The best and second-best results are highlighted in \textcolor{red}{red} and \textcolor{blue}{blue}, respectively.}
\label{tab:Quantitative_Evaluation}
\resizebox{\textwidth}{!}{%
\begin{tabular}{l|c|ccccc|cccc}
\toprule
\multicolumn{1}{c|}{\multirow{2}{*}{Method}} & \multirow{2}{*}{Type} & \multicolumn{5}{c|}{FFHQ} & \multicolumn{4}{c}{WildDark-Face} \\ 
\cline{3-11}
\multicolumn{1}{c|}{} & & PSNR$\uparrow$ & LPIPS$\downarrow$ & DISTS$\downarrow$ & Deg.$\downarrow$ & LMD$\downarrow$ & MUSIQ$\uparrow$ & MANIQA$\uparrow$ & HyperIQA$\uparrow$ & FID$\downarrow$ \\ 
\midrule

LightenDiffusion~\cite{jiang2024lightendiffusion} $\rightarrow$ TSD-SR~\cite{dong2025tsd} & C & 17.85 & 0.4126 & 0.2745 & 9.0737 & 2.6448 & 42.5937 & 0.2644 & 0.5817 & 191.74 \\
LightenDiffusion~\cite{jiang2024lightendiffusion} $\rightarrow$ DiffBIR~\cite{lin2024diffbir} & C & 18.20 & \textcolor{blue}{0.3063} & \textcolor{blue}{0.2284} & \textcolor{blue}{7.2990} & \textcolor{blue}{2.1906} & \textcolor{blue}{49.4678} & \textcolor{blue}{0.3302} & \textcolor{blue}{0.6335} & \textcolor{red}{121.47} \\
TSD-SR~\cite{dong2025tsd} $\rightarrow$ LightenDiffusion~\cite{jiang2024lightendiffusion} & C & 17.52 & 0.3850 & 0.2931 & 9.0773 & 3.0751 & 24.3632 & 0.1855 & 0.2680 & 157.44 \\
DiffBIR~\cite{lin2024diffbir} $\rightarrow$ LightenDiffusion~\cite{jiang2024lightendiffusion} & C & 16.62 & 0.4591 & 0.3024 & 8.9765 & 2.8251 & 49.1503 & 0.2879 & 0.5748 & 177.81 \\
\hline

DarkIR~\cite{feijoo2025darkir} & J & 15.74 & 0.4487 & 0.3090 & 9.1225 & 2.6302 & 18.5099 & 0.1665 & 0.2131 & 186.82 \\
LIEDNet~\cite{liu2025liednet} & J & 15.50 & 0.4397 & 0.3084 & 9.1932 & 2.6580 & 18.6445 & 0.1645 & 0.2116 & 183.96 \\
LEDNet~\cite{zhou2022lednet} & J & 16.49 & 0.4135 & 0.2956 & 8.9023 & 2.8506 & 21.1880 & 0.1830 & 0.2277 & 173.70 \\
VQCNIR~\cite{zou2024vqcnir} & J & 14.35 & 0.4618 & 0.3164 & 9.3756 & 2.7004 & 18.8857 & 0.1870 & 0.2264 & 177.35 \\
URWKV~\cite{xu2025urwkv} & J & 13.59 & 0.4747 & 0.3321 & 9.5086 & 2.7045 & 18.5134 & 0.1671 & 0.2052 & 176.95 \\
FDN~\cite{tu2025fourier} & J & \textcolor{blue}{18.56} & 0.4428 & 0.3031 & 9.4073 & 2.3465 & 16.2443 & 0.1451 & 0.2036 & 161.70 \\
\hline

\textbf{PASDiff (Ours)} & J & \textcolor{red}{22.59} & \textcolor{red}{0.2559} & \textcolor{red}{0.2274} & \textcolor{red}{6.8031} & \textcolor{red}{1.6213} & \textcolor{red}{53.1825} & \textcolor{red}{0.3658} & \textcolor{red}{0.7057} & \textcolor{blue}{127.04} \\

\bottomrule
\end{tabular}%
}
\end{table*}

\noindent\textbf{Datasets and Metrics.} 
For synthetic test datasets, we utilize 1,000 high-quality images from FFHQ, uniformly resized to $256 \times 256$. We then apply a physically grounded two-stage degradation pipeline to approximate real-world low-light conditions. Following the protocol in~\cite{li2020blind, li2018learning}, the first stage simulates general face degradation: the ground truth $y$ is sequentially degraded by Gaussian blur ($\sigma \in [0.1, 5]$), downsampling ($r \in [1, 4]$), additive white Gaussian noise ($\delta \in [0, 15]$), and JPEG compression ($q \in [60, 100]$). The second stage simulates low-light physics in the linear domain: we apply an exposure adjustment factor $\alpha = 0.25$ and a hue-preserving Gamma correction sampled from $\gamma \in [1.7, 1.9]$. Formally, the overall degradation process is formulated as:
\begin{equation}
    x = \text{Gamma}_{\gamma}\Big( \alpha \cdot \text{JPEG}_{q}\big( (y \circledast k_{\sigma}) \downarrow_{r} + \mathbf{n}_{\delta} \big) \Big),
\end{equation}
where $x$ denotes the synthesized low-light facial image, $\alpha$ is the exposure adjustment factor, $k_{\sigma}$ is the Gaussian blur kernel, $\downarrow_{r}$ is the downsampling operation, and $\mathbf{n}_{\delta}$ represents the additive white Gaussian noise.

To further assess real-world scenes, we construct WildDark-Face, a real-world benchmark comprising 700 in-the-wild low-light facial images. Specifically, we derive this dataset from the widely recognized DarkFace benchmark~\cite{poor_visibility_benchmark} by cropping individual faces using the provided bounding box annotations. To ensure evaluation reliability, we filter out instances with extremely low spatial resolutions and meticulously curate a representative subset. These real-world captures encapsulate a wide spectrum of authentic, unconstrained degradations, encompassing extreme photon starvation, severe sensor noise, unpredictable motion blur, and diverse head poses. For quantitative evaluation on synthetic data, we employ PSNR and LPIPS~\cite{zhang2018unreasonable} for signal and perceptual fidelity, and DISTS~\cite{ding2020image} for texture similarity. For real-world assessment, we utilize no-reference metrics including MUSIQ~\cite{ke2021musiq}, MANIQA~\cite{yang2022maniqa}, HyperIQA~\cite{su2020blindly}, and FID~\cite{heusel2017gans}. Furthermore, we report Deg. (degree of identity preservation)~\cite{deng2019arcface} and LMD (Landmark Distance)~\cite{wang2019adaptive} to explicitly measure semantic and geometric identity alignment. More details about the WildDark-Face dataset are provided in the \href{https://guangweigao.github.io/paper/26-ECCV-Supplements.pdf}{\textcolor{purple}{Supplements}}.
% \footnote{https://guangweigao.github.io/paper/Supplements.pdf}

\begin{figure*}[t]
    \centering
    \setlength{\tabcolsep}{0pt} 

    \includegraphics[width=0.125\linewidth]{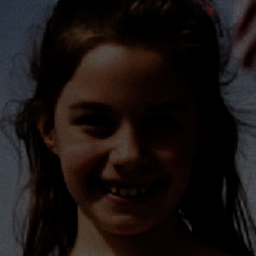}\hfill
    \includegraphics[width=0.125\linewidth]{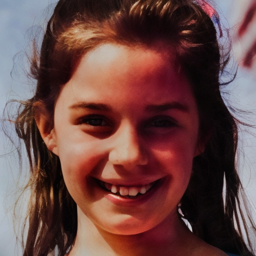}\hfill
    \includegraphics[width=0.125\linewidth]{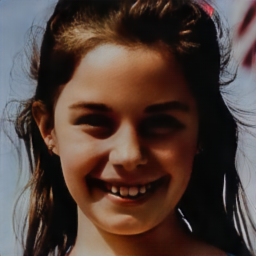}\hfill
    \includegraphics[width=0.125\linewidth]{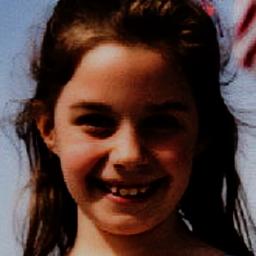}\hfill
    \includegraphics[width=0.125\linewidth]{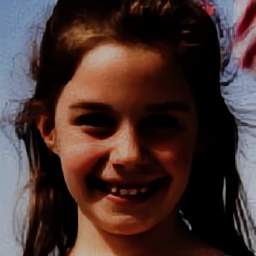}\hfill
    \includegraphics[width=0.125\linewidth]{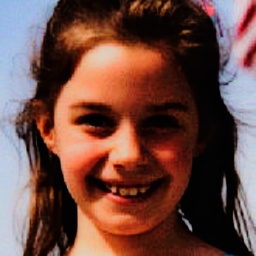}\hfill
    \includegraphics[width=0.125\linewidth]{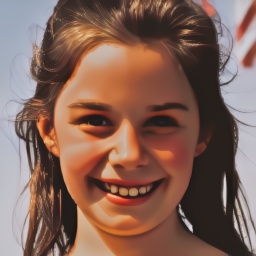}\hfill
    \includegraphics[width=0.125\linewidth]{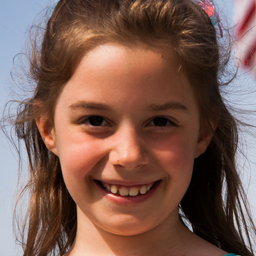}\\
    \vspace{1pt}

    \includegraphics[width=0.125\linewidth]{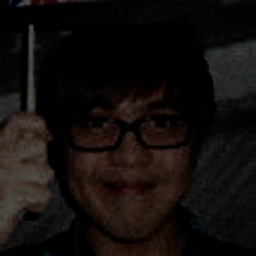}\hfill
    \includegraphics[width=0.125\linewidth]{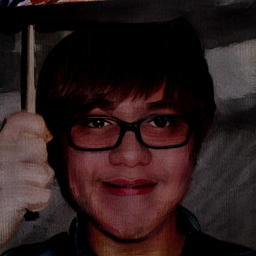}\hfill
    \includegraphics[width=0.125\linewidth]{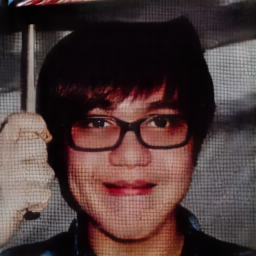}\hfill
    \includegraphics[width=0.125\linewidth]{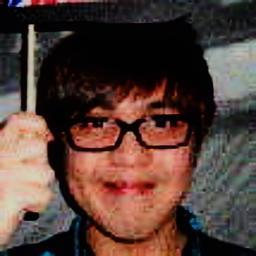}\hfill
    \includegraphics[width=0.125\linewidth]{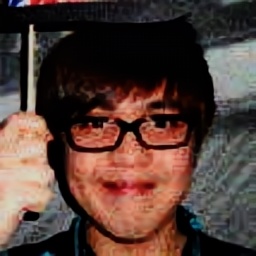}\hfill
    \includegraphics[width=0.125\linewidth]{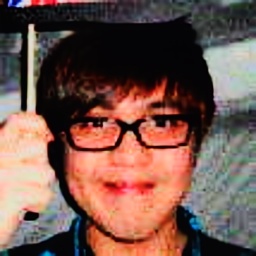}\hfill
    \includegraphics[width=0.125\linewidth]{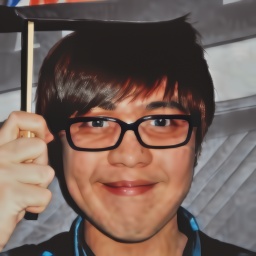}\hfill
    \includegraphics[width=0.125\linewidth]{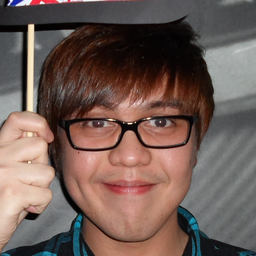}\\
    \vspace{1pt}

    \includegraphics[width=0.125\linewidth]{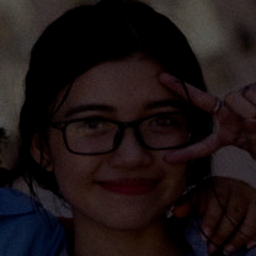}\hfill
    \includegraphics[width=0.125\linewidth]{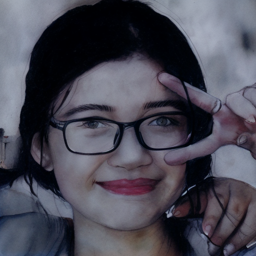}\hfill
    \includegraphics[width=0.125\linewidth]{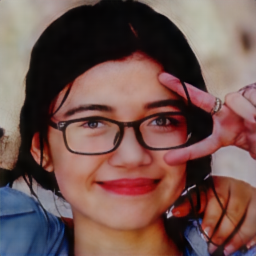}\hfill
    \includegraphics[width=0.125\linewidth]{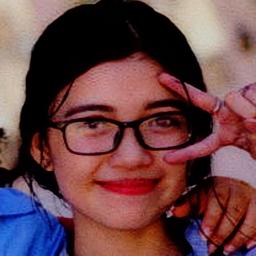}\hfill
    \includegraphics[width=0.125\linewidth]{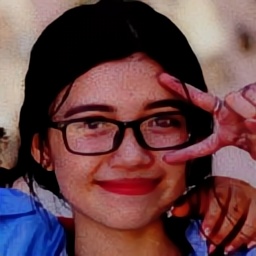}\hfill
    \includegraphics[width=0.125\linewidth]{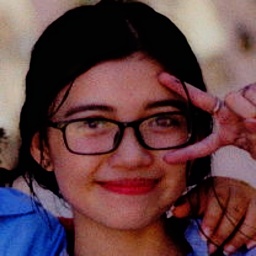}\hfill
    \includegraphics[width=0.125\linewidth]{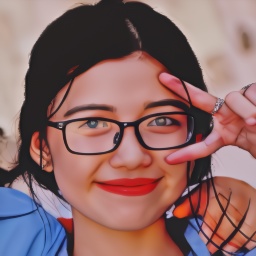}\hfill
    \includegraphics[width=0.125\linewidth]{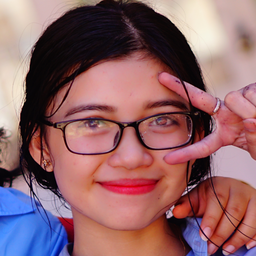}\\
    \vspace{2pt}

    \begin{minipage}[t]{0.125\linewidth} \centering \fontsize{5.5pt}{6pt}\selectfont LQ \end{minipage}\hfill
    \begin{minipage}[t]{0.125\linewidth} \centering \fontsize{5.5pt}{6pt}\selectfont L-Diff$\to$\\DiffBIR \end{minipage}\hfill
    \begin{minipage}[t]{0.125\linewidth} \centering \fontsize{5.5pt}{6pt}\selectfont DiffBIR\\$\to$L-Diff \end{minipage}\hfill
    \begin{minipage}[t]{0.125\linewidth} \centering \fontsize{5.5pt}{6pt}\selectfont DarkIR \end{minipage}\hfill
    \begin{minipage}[t]{0.125\linewidth} \centering \fontsize{5.5pt}{6pt}\selectfont LEDNet \end{minipage}\hfill
    \begin{minipage}[t]{0.125\linewidth} \centering \fontsize{5.5pt}{6pt}\selectfont FDN \end{minipage}\hfill
    \begin{minipage}[t]{0.125\linewidth} \centering \fontsize{5.5pt}{6pt}\selectfont \textbf{Ours} \end{minipage}\hfill
    \begin{minipage}[t]{0.125\linewidth} \centering \fontsize{5.5pt}{6pt}\selectfont GT \end{minipage}
    \caption{Visual comparisons on the synthetic FFHQ dataset.}
    \label{fig:visual_ffhq}
\end{figure*}

\subsection{Comparisons with Existing Methods}

Since existing methods typically treat low-light enhancement and face restoration in isolation, there is currently no end-to-end solution specifically tailored for this joint task. To provide an evaluation, we compare PASDiff against two categories. First, we examine Cascaded Approaches by combining low-light enhancers with blind face restorers. Specifically, we employ LightenDiffusion~\cite{jiang2024lightendiffusion} as the enhancer, and select TSD-SR~\cite{dong2025tsd} and DiffBIR~\cite{lin2024diffbir} as restorers, evaluating both Enhancement $\rightarrow$ Restoration and Restoration $\rightarrow$ Enhancement cascades. Second, we compare against Joint General Restoration Approaches designed for low-light enhancement and restoration in generic scenes, including DarkIR~\cite{feijoo2025darkir}, LIEDNet~\cite{liu2025liednet}, LEDNet~\cite{zhou2022lednet}, VQCNIR~\cite{zou2024vqcnir}, URWKV~\cite{xu2025urwkv}, and FDN~\cite{tu2025fourier}.

\begin{figure*}[t]
    \centering
    \setlength{\tabcolsep}{0pt} 

     \includegraphics[width=0.142\linewidth]{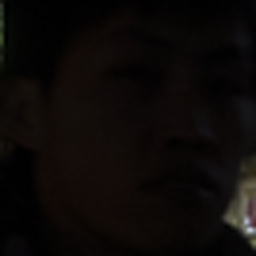}\hfill
    \includegraphics[width=0.142\linewidth]{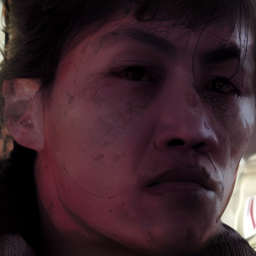}\hfill
    \includegraphics[width=0.142\linewidth]{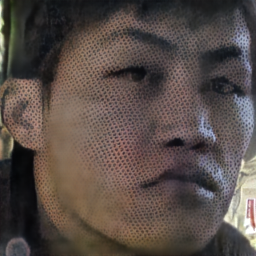}\hfill
    \includegraphics[width=0.142\linewidth]{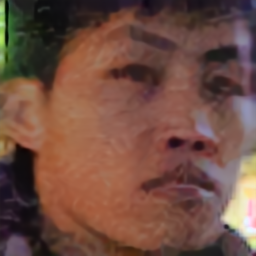}\hfill
    \includegraphics[width=0.142\linewidth]{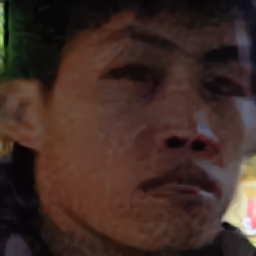}\hfill
    \includegraphics[width=0.142\linewidth]{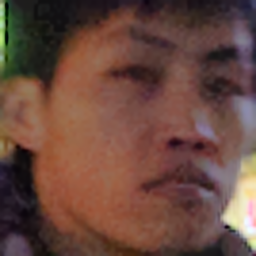}\hfill
    \includegraphics[width=0.142\linewidth]{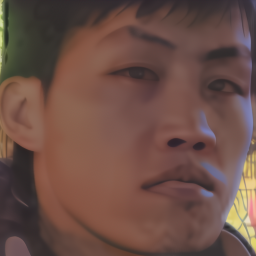}\\
    \vspace{1pt}

    \includegraphics[width=0.142\linewidth]{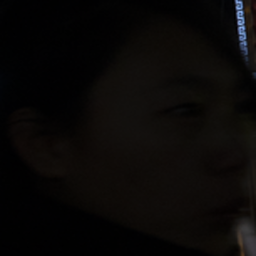}\hfill
    \includegraphics[width=0.142\linewidth]{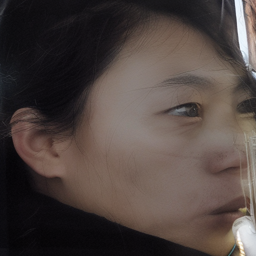}\hfill
    \includegraphics[width=0.142\linewidth]{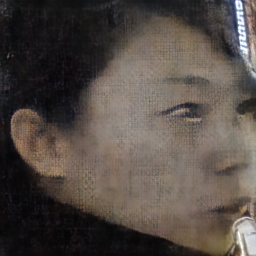}\hfill
    \includegraphics[width=0.142\linewidth]{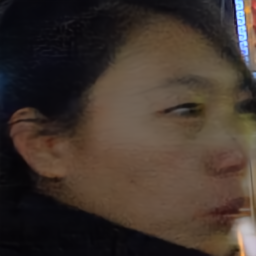}\hfill
    \includegraphics[width=0.142\linewidth]{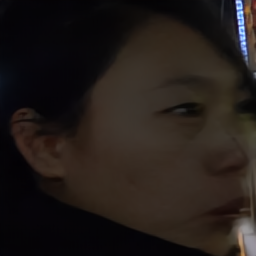}\hfill
    \includegraphics[width=0.142\linewidth]{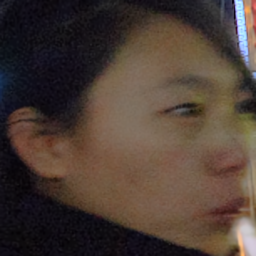}\hfill
    \includegraphics[width=0.142\linewidth]{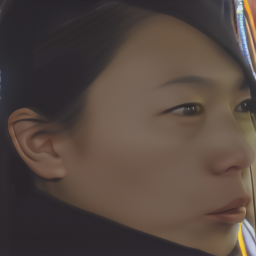}\\
    \vspace{1pt}

    \includegraphics[width=0.142\linewidth]{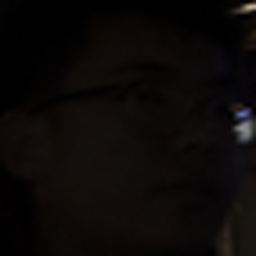}\hfill
    \includegraphics[width=0.142\linewidth]{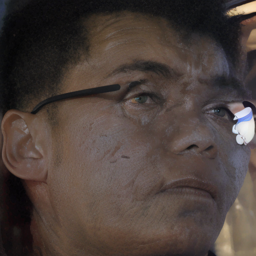}\hfill
    \includegraphics[width=0.142\linewidth]{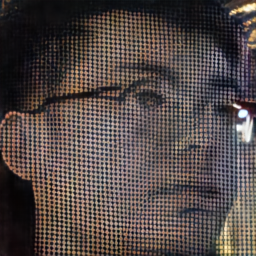}\hfill
    \includegraphics[width=0.142\linewidth]{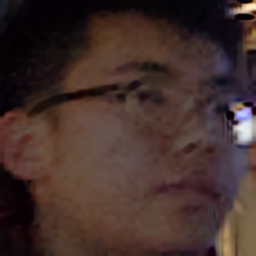}\hfill
    \includegraphics[width=0.142\linewidth]{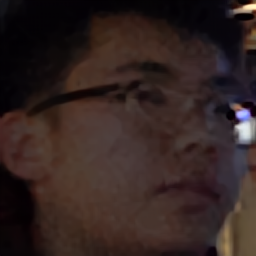}\hfill
    \includegraphics[width=0.142\linewidth]{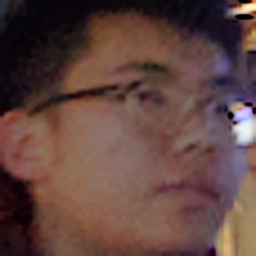}\hfill
    \includegraphics[width=0.142\linewidth]{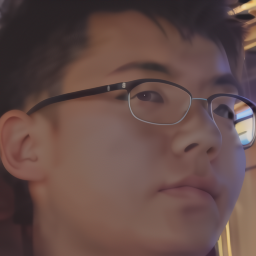}\\
    \vspace{2pt}

    \begin{minipage}[t]{0.142\linewidth} \centering \fontsize{5.5pt}{6pt}\selectfont LQ \end{minipage}\hfill
    \begin{minipage}[t]{0.142\linewidth} \centering \fontsize{5.5pt}{6pt}\selectfont L-Diff$\to$\\DiffBIR \end{minipage}\hfill
    \begin{minipage}[t]{0.142\linewidth} \centering \fontsize{5.5pt}{6pt}\selectfont DiffBIR\\$\to$L-Diff \end{minipage}\hfill
    \begin{minipage}[t]{0.142\linewidth} \centering \fontsize{5.5pt}{6pt}\selectfont DarkIR \end{minipage}\hfill
    \begin{minipage}[t]{0.142\linewidth} \centering \fontsize{5.5pt}{6pt}\selectfont LEDNet \end{minipage}\hfill
    \begin{minipage}[t]{0.142\linewidth} \centering \fontsize{5.5pt}{6pt}\selectfont FDN \end{minipage}\hfill
    \begin{minipage}[t]{0.142\linewidth} \centering \fontsize{5.5pt}{6pt}\selectfont \textbf{Ours} \end{minipage}
    \caption{Visual comparisons on the real-world WildDark-Face dataset.}
    \label{fig:visual_WildDark-Face}
\end{figure*}

\noindent\textbf{Quantitative Evaluation.} 
Table~\ref{tab:Quantitative_Evaluation} summarizes quantitative results on both synthetic and real test sets. While the cascaded combination of LightenDiffusion~\cite{jiang2024lightendiffusion} and DiffBIR~\cite{lin2024diffbir} achieves competitive perceptual scores, it suffers from inherent errors accumulation typical of multi-stage pipelines. Similarly, although generic restoration models such as FDN achieve decent PSNR scores, they fail to capture high-frequency facial semantics, resulting in poor performance on identity-aware metrics (Deg. and LMD). In contrast, PASDiff achieves the best performance across almost all evaluation metrics.

Furthermore, we conduct a face recognition accuracy test using the pre-trained InsightFace (\texttt{buffalo\_l}) model~\cite{deng2019arcface}. Specifically, we extract normalized facial embeddings from both restored and ground-truth faces to compute cosine similarity, defining a successful identity match with a threshold of 0.42. As reported in Table~\ref{tab:face_acc}, our method achieves over 8\% higher face recognition accuracy than the second-best method.

\noindent\textbf{Qualitative Evaluation.}
Visual comparisons on the synthetic FFHQ and real-world WildDark-Face datasets are presented in Fig.~\ref{fig:visual_ffhq} and Fig.~\ref{fig:visual_WildDark-Face}, respectively. As observed, cascaded methods exhibit specific limitations depending on their execution order. In the L-Diff~\cite{jiang2024lightendiffusion}$\to$DiffBIR~\cite{lin2024diffbir} sequence, the restoration model tends to hallucinate unnatural textures based on noise amplified by the preceding enhancer, resulting in blotchy color artifacts and structural fractures on faces. Conversely, the reverse order (DiffBIR~\cite{lin2024diffbir}$\to$L-Diff~\cite{jiang2024lightendiffusion}) often suffers from grid-like artifacts and severe detail degradation. Meanwhile, representative joint general methods (e.g., LEDNet, FDN, and DarkIR) struggle with such complex composite degradations; they fail to effectively lift the illumination and frequently produce excessively blurry and under-enhanced results. In contrast, PASDiff generates photometrically natural and identity-consistent facial features. More visual results of other baseline methods 
are provided in the \href{https://guangweigao.github.io/paper/26-ECCV-Supplements.pdf}{\textcolor{purple}{Supplements}}. %More qualitative evaluations can be found in the \href{https://guangweigao.github.io/paper/Supplements.pdf}{\textcolor{purple}{Supplements}}.

\noindent\textbf{User Study.} 
To capture human visual preference, we conducted a subjective user study using 10 synthetic and 10 real-world images. Evaluators assessed our method against the top three baselines by ranking the results across three dimensions: natural lighting, chromatic realism, and structural clarity. As illustrated in Fig.~\ref{fig:priors_user} (Right), PASDiff consistently receives the highest preference rankings, validating its superiority in generating visually pleasing results.

\subsection{Ablation Study}

 To further validate the effectiveness of each component, we report ablation results on our synthetic dataset concerning the physical constraints, structural injection, and style-agnostic design.

\begin{figure*}[t]
    \centering
    \begin{minipage}[c]{0.38\linewidth}
        \makeatletter\def\@captype{table}\makeatother 
        \centering
        \caption{Ablation on physical constraints.}
        \label{tab:exp_ref}
        \resizebox{\linewidth}{!}{%
        \begin{tabular}{lccc}
        \toprule
        \textbf{Method} & \textbf{PSNR}$\uparrow$ & \textbf{LPIPS}$\downarrow$ & \textbf{LMD}$\downarrow$ \\
        \midrule
        w/o $\mathcal{L}_{exp}$ & 21.54 & \textbf{0.2463} & 1.6295 \\ 
        w/o $\mathcal{L}_{ref}$ & 11.15 & 0.6680 & 13.3718 \\
        \midrule
        \textbf{Ours} & \textbf{22.59} & 0.2559 & \textbf{1.6213} \\ 
        \bottomrule
        \end{tabular}%
        }
    \end{minipage}\hfill
    \begin{minipage}[c]{0.56\linewidth}
        \makeatletter\def\@captype{figure}\makeatother 
        \centering
        \includegraphics[width=0.235\linewidth]{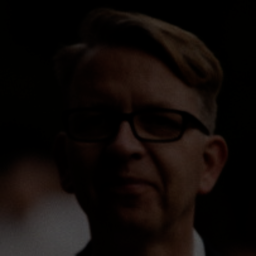}\hfill
        \includegraphics[width=0.235\linewidth]{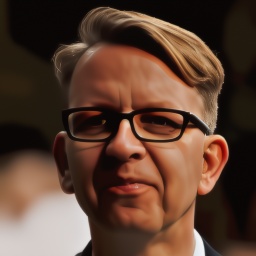}\hfill
        \includegraphics[width=0.235\linewidth]{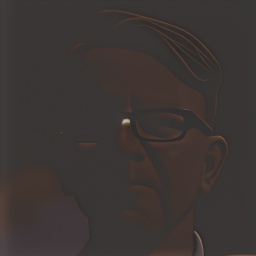}\hfill
        \includegraphics[width=0.235\linewidth]{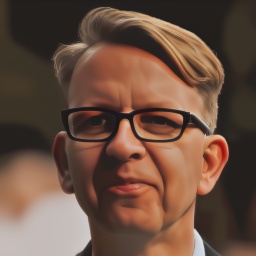}\\
        \vspace{1pt}
        \begin{minipage}[t]{0.235\linewidth} \centering \fontsize{6pt}{7pt}\selectfont LQ \end{minipage}\hfill
        \begin{minipage}[t]{0.235\linewidth} \centering \fontsize{6pt}{7pt}\selectfont w/o $\mathcal{L}_{exp}$ \end{minipage}\hfill
        \begin{minipage}[t]{0.235\linewidth} \centering \fontsize{6pt}{7pt}\selectfont w/o $\mathcal{L}_{ref}$ \end{minipage}\hfill
        \begin{minipage}[t]{0.235\linewidth} \centering \fontsize{6pt}{7pt}\selectfont \textbf{Ours} \end{minipage}
        \caption{Visual ablation on physical constraints.}
        \label{fig:exp_ref}
    \end{minipage}
\end{figure*}

\begin{figure*}[t]
    \centering
    \begin{minipage}[c]{0.38\linewidth}
        \makeatletter\def\@captype{table}\makeatother 
        \centering
        \caption{Ablation on structural injection.}
        \label{tab:stru}
        \resizebox{\linewidth}{!}{%
        \begin{tabular}{lccc}
        \toprule
        \textbf{Method} & \textbf{PSNR}$\uparrow$ & \textbf{LPIPS}$\downarrow$ & \textbf{LMD}$\downarrow$ \\
        \midrule
        w/o $\mathcal{L}_{stru}$ & \textbf{23.57} & 0.4501 & 2.1251 \\ 
        \midrule
        \textbf{Ours} & 22.59 & \textbf{0.2559} & \textbf{1.6213} \\ 
        \bottomrule
        \end{tabular}%
        }
    \end{minipage}\hfill
    \begin{minipage}[c]{0.56\linewidth}
        \makeatletter\def\@captype{figure}\makeatother 
        \centering
        \includegraphics[width=0.235\linewidth]{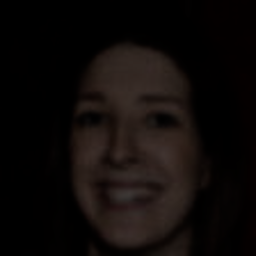}\hspace{0.02\linewidth}%
        \includegraphics[width=0.235\linewidth]{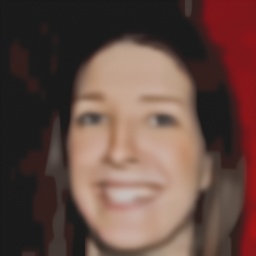}\hspace{0.02\linewidth}%
        \includegraphics[width=0.235\linewidth]{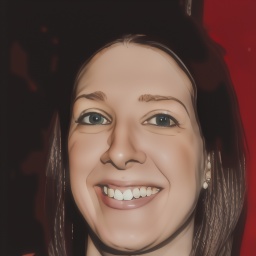}\\
        \vspace{1pt}
        \begin{minipage}[t]{0.235\linewidth} \centering \fontsize{6pt}{7pt}\selectfont LQ \end{minipage}\hspace{0.02\linewidth}%
        \begin{minipage}[t]{0.235\linewidth} \centering \fontsize{6pt}{7pt}\selectfont w/o $\mathcal{L}_{stru}$ \end{minipage}\hspace{0.02\linewidth}%
        \begin{minipage}[t]{0.235\linewidth} \centering \fontsize{6pt}{7pt}\selectfont \textbf{Ours} \end{minipage}
        \caption{Visual ablation on structural injection.}
        \label{fig:stru}
    \end{minipage}
\end{figure*}

\noindent\textbf{Impact of Physical Constraints.} 
The physical constraints, comprising the exposure loss $\mathcal{L}_{exp}$ and the reflectance loss $\mathcal{L}_{ref}$, form the foundation of our restoration. As shown in Table~\ref{tab:exp_ref} and Fig.~\ref{fig:exp_ref}, when removing $\mathcal{L}_{exp}$, although the basic content remains visible due to the reflectance constraint, the restored images appear overall darker, failing to meet the visibility standards of normal-light scenes. Conversely, removing $\mathcal{L}_{ref}$ deprives the model of a reliable chromatic anchor. The results suffer from a profound loss of chromatic information, failing to recover their intrinsic colors. This confirms that both terms are essential for establishing a physically plausible foundation.

\noindent\textbf{Impact of Structural Injection.} 
Removing the structural guidance loss $\mathcal{L}_{stru}$ forces the model to rely solely on physical constraints. As illustrated in Table~\ref{tab:stru} and Fig.~\ref{fig:stru}, the resulting images suffer from residual blur, losing critical high-frequency identity characteristics. Although this low-frequency-biased optimization trivially yields a higher PSNR, it severely degrades perceptual fidelity (LPIPS) and structural alignment (LMD), which demonstrates that physics-based guidance alone is insufficient for realistic blind face restoration.

\begin{figure*}[t]
    \centering
    \begin{minipage}[c]{0.38\linewidth}
        \makeatletter\def\@captype{table}\makeatother 
        \centering
        \caption{Ablation on style-agnostic design.}
        \label{tab:sasi}
        \resizebox{\linewidth}{!}{%
        \begin{tabular}{lccc}
        \toprule
        \textbf{Method} & \textbf{PSNR}$\uparrow$ & \textbf{LPIPS}$\downarrow$ & \textbf{LMD}$\downarrow$ \\
        \midrule
        MSE & 21.86 & 0.2571 & 1.6518 \\
        \midrule
        \textbf{Ours} & \textbf{22.59} & \textbf{0.2559} & \textbf{1.6213} \\
        \bottomrule
        \end{tabular}%
        }
    \end{minipage}\hfill
    \begin{minipage}[c]{0.56\linewidth}
        \makeatletter\def\@captype{figure}\makeatother 
        \centering
        \includegraphics[width=0.235\linewidth]{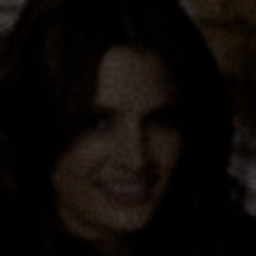}\hfill
        \includegraphics[width=0.235\linewidth]{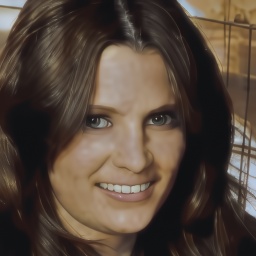}\hfill
        \includegraphics[width=0.235\linewidth]{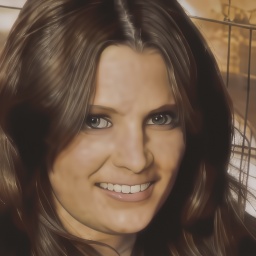}\hfill
        \includegraphics[width=0.235\linewidth]{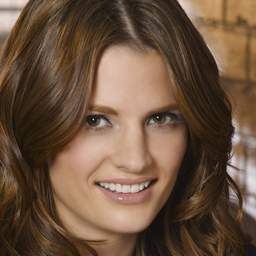}\\
        \vspace{1pt}
        \begin{minipage}[t]{0.235\linewidth} \centering \fontsize{6pt}{7pt}\selectfont LQ \end{minipage}\hfill
        \begin{minipage}[t]{0.235\linewidth} \centering \fontsize{6pt}{7pt}\selectfont MSE \end{minipage}\hfill
        \begin{minipage}[t]{0.235\linewidth} \centering \fontsize{6pt}{7pt}\selectfont \textbf{Ours} \end{minipage}\hfill
        \begin{minipage}[t]{0.235\linewidth} \centering \fontsize{6pt}{7pt}\selectfont GT \end{minipage}
        \caption{Visual ablation on style-agnostic design.}
        \label{fig:sasi}
    \end{minipage}
\end{figure*}

\noindent\textbf{Effectiveness of Style-Agnostic Design.} 
We evaluate the necessity of our SASI by replacing it with an MSE loss between estimated images and the prior's output. Results in Table~\ref{tab:sasi} and Fig.~\ref{fig:sasi} reveal a critical gradient conflict: the naive approach forces diffusion models to mimic the prior's synthetic lighting style, directly contradicting our physically grounded exposure and reflectance constraints. Consequently, this mismatch leads to distinct color shifts and discordant global tones. In contrast, our AdaIN-based SASI successfully distills structural semantics while statistically filtering out mismatched styles, effectively harmonizing high-fidelity details with the correct physical atmosphere.

\section{Limitations and Future Work} 
Despite achieving SOTA quality, PASDiff presents two main limitations. First, relying on iterative diffusion sampling incurs a slower inference speed compared to end-to-end feed-forward networks. Second, recovering precise chromaticity from extremely low-light inputs remains intrinsically ill-posed. Although our physical constraints significantly shift the overall color distribution toward high-quality references, the model can still exhibit a certain degree of under-saturation in near-pitch-black regions where the original color information is irreversibly destroyed. In future work, we intend to integrate advanced diffusion acceleration techniques to streamline inference and explore explicit generative color priors to further close the chromaticity gap in extreme real degradations.

\section{Conclusion}

We propose PASDiff, a training-free framework tailored for joint low-light enhancement and blind face restoration. To achieve physically plausible illumination and color distributions, we introduce photometric constraints leveraging inverse intensity weighting and Retinex theory. To faithfully recover high-frequency facial details, we devise a SASI strategy, which distills structural semantics from an off-the-shelf facial prior while explicitly filtering out its intrinsic photometric biases. By seamlessly harmonizing physical and semantic constraints, PASDiff effectively resolves the error accumulation and structural infidelity issues prevalent in existing cascaded and joint paradigms. Extensive evaluations on both synthetic and real-world benchmarks demonstrate the superiority of our approach in yielding naturally illuminated, chromatically plausible, and identity-consistent.

\section*{Acknowledgements}
This work was supported in part by the National Natural Science Foundation of China under Grant Nos. U24A20330 and 62361166670.

\bibliographystyle{splncs04}
\bibliography{main}

\end{document}